\documentclass[letterpaper]{article} % DO NOT CHANGE THIS
\usepackage{aaai2026}  % DO NOT CHANGE THIS
\usepackage{times}  % DO NOT CHANGE THIS
\usepackage{helvet}  % DO NOT CHANGE THIS
\usepackage{courier}  % DO NOT CHANGE THIS
\usepackage[hyphens]{url}  % DO NOT CHANGE THIS
\usepackage{graphicx} % DO NOT CHANGE THIS
\usepackage{natbib}  % DO NOT CHANGE THIS AND DO NOT ADD ANY OPTIONS TO IT
\usepackage{caption} % DO NOT CHANGE THIS AND DO NOT ADD ANY OPTIONS TO IT
\usepackage{comment}
\usepackage{algorithm}
\usepackage{algorithmic}
\usepackage{amssymb}
\usepackage{amsmath}

\usepackage{subcaption}
\usepackage{booktabs}

\usepackage{listings}
\floatstyle{ruled}
\newfloat{listing}{tb}{lst}{}
\floatname{listing}{Listing}
\title{Petri Net Induced Heuristic Search\\ for Resource Constrained Scheduling}
\author{
    Ido Lublin,
    Dor Atzmon,
    Izack Cohen
}
\affiliations{
    Faculty of Engineering\\
    Bar-Ilan University\\
    Ramat Gan, Israel\\
    \{ido.lublin, dor.atzmon, izack.cohen\}@biu.ac.il
}
\usepackage{amsthm}

\newtheorem{example}{Example}%}

\begin{document}

\maketitle

\begin{abstract}
We formulate the \emph{Resource-Constrained Project Scheduling Problem} (RCPSP) as optimal search over the reachability graph of a \emph{Timed Transition Petri Net with Resources}, using relative-delay tokens so that scheduling decisions correspond to transition firings in the induced state space. We solve the resulting problem with $A^*$ guided by a heuristic that combines Critical Path and resource-based lower bounds, and prove that it is consistent under our token-based time semantics. Experiments on the PSPLIB benchmarks show that the approach outperforms strong exact Mixed-Integer Linear Programming (MIP) baselines (SCIP, CBC) in both success rate and solve time. Per-instance analysis shows that heuristic search and MIP degrade along independent axes, resource tightness for $A^*$ and formulation size for MIP, with resource strength mediating which solver benefits from scale.
\end{abstract}


\section{Introduction}

The \emph{Resource-Constrained Project Scheduling Problem} (RCPSP) asks how to schedule precedence-constrained activities under limited shared resources so as to minimize completion time~\cite{herroelen1998resource, brucker1999resource, hartmann2022updated}. As a strongly NP-hard generalization of classical scheduling problems~\cite{BLAZEWICZ198311}, RCPSP is a fundamental problem at the intersection of scheduling, combinatorial search, and resource-bounded reasoning. It also arises in temporal planning, robotic task coordination, spacecraft mission scheduling, multiprocessor scheduling, and project management. RCPSP has long served as a benchmark for exact and approximate methods, yet no single method dominates across all instance classes.

The leading exact solution approach is \emph{Mixed-Integer Linear Programming} (MIP), typically solved by branch-and-cut~\cite{dorndorf2000branch, kone2011event, artigues2015mixed}. While MIP provides optimality guarantees, it scales poorly, and strong solvers such as SCIP~\cite{achterberg2009scip} often fail to solve benchmark instances within practical time limits. Metaheuristics~\cite{pellerin2020survey} improve scalability at the cost of optimality, and hybrid methods seek a middle ground, yet no existing approach reliably achieves both. A natural exact alternative is \emph{heuristic search}~\cite{hart1968formal}. ~\citet{repec:wly:navres:v:37:y:1990:i:1:p:61-84} applied $A^*$ to RCPSP through conflict sequencing, but the method becomes intractable on highly resource-constrained instances. ~\citet{Zamani01071998} introduced the activity-dispatch state space and applied $A^*$ via \emph{SLA*}, improving performance but requiring multiple search trials to reach optimality. In both cases, the lack of a consistent heuristic prevents safe duplicate pruning and causes redundant re-expansion of equivalent scheduling states--we tackle this issue in the current work. More broadly, a recent survey~\cite{huang2023scheduling} highlights the promise of heuristic search over reachability graphs for balancing solution quality and computational effort. Related Petri-net-based ideas have been explored in project and manufacturing scheduling~\cite{mejia2016petri, mejia2005approach}, directed heuristic search over Petri net reachability graphs has 
been studied in the model checking context~\cite{edelkamp2006action}, and simulation has been used to estimate RCPSP project duration~\cite{wu2009solving}. Yet, to the best of our knowledge, no prior work has applied optimal heuristic search to the induced reachability graph of an RCPSP formulation.

This paper develops a new RCPSP formulation as an optimal search over the reachability graph induced by a \emph{Timed Transition Petri Net with Resources} (TTPNR), making local scheduling decisions explicit and representing time uniformly through token delays. This yields an exact search formulation with admissible heuristics, safe duplicate pruning, and efficient treatment of zero-cost transitions. We prove that the combined Critical Path and resource-based heuristic is consistent, and show on standard PSPLIB benchmarks that the resulting approach is competitive with strong exact MIP baselines, including SCIP and CBC.

\iffalse
The contributions of this work are:
\begin{enumerate}
    \item A new formulation of RCPSP as a reachability problem 
    over a Timed Transition Petri Net with relative-delay token 
    encoding, transforming scheduling into a pathfinding problem 
    amenable to heuristic search.
    \item A formal proof that the CP heuristic is consistent under 
    this formulation, justifying the closed list and enabling a 
    caching optimization for zero-cost transitions unique to our 
    state representation.
    \item An empirical evaluation on standard PSPLIB benchmarks, demonstrating performance competitive 
    with state-of-the-art MIP solvers SCIP and CBC.
\end{enumerate}
\fi

\section{RCPSP Problem Definition}

RCPSP is characterized by the tuple $(\mathbb{A}, E, R, C, U, \tau)$:

\begin{itemize}
    \item $\mathbb{A} = \{0, 1, \dots, n, n+1\}$: set of activities, where $0$ and $n+1$ are dummy start and finish.
    \item $E \subset \mathbb{A} \times \mathbb{A}$: precedence constraints, 
    where $(i, j) \in E$ means $j$ cannot start before $i$ completes. 
    Let $\bullet j = \{i \mid (i,j) \in E\}$ be the set of 
    immediate predecessors of $j$.
    \item $R = \{r_1, \dots, r_k\}$: renewable resource types with capacities $C = \{c_1, \dots, c_k\}$.
    \item $u_{j,i}$: demand of activity $j$ for resource $r_i$.
    \item $\tau_j$: duration of activity $j$.
\end{itemize}
As in standard RCPSP formulations, the dummy activities have zero duration
($\tau_0=\tau_{n+1}=0$) and zero resource demand
($u_{0,i}=u_{n+1,i}=0$ for all $r_i \in R$).

A project \emph{schedule} assigns a start time $S_j \ge 0$ to each activity $j \in \mathbb{A}$.
A feasible schedule satisfies two constraints:

\begin{enumerate}
    \item \emph{Precedence:} An activity cannot start until all of its predecessors complete (for every $(i,j) \in E$,
    $
        S_j \ge S_i + \tau_i
    $).

    \item \emph{Resource:} At any time $t$, %during the project execution, 
    the total amount of each resource type $r_i \in R$ allocated to ongoing activities cannot exceed its available capacity $c_i \in C$. %Formally, letting $\mathcal{O}(t) \subseteq \mathcal{A}$ denote the set of activities in execution at time $t$, we require
    %$
    %    \sum_{j \in \mathcal{O}(t)} u_{j,i} \le c_i, %\qquad 
    %    \forall r_i \in R.
    %$
\end{enumerate}

The objective is to find a schedule with minimal makespan, that is, the start time of the dummy finish activity:
$
\min S_{n+1}.
$
The interaction between precedence and resource constraints creates a highly constrained combinatorial problem, which we later formulate both as a reachability graph and as a MIP. 
We use the next example throughout the paper.

\begin{example}\label{exm1}
Consider an RCPSP instance with activities
$
\mathbb{A}=\{0,1,2,3,4,5\}
$
(denote by
$
0\equiv a,\;1\equiv b,\;2\equiv c,\;$ etc.). $a$ and $f$ are the dummy start and finish activities.
The precedence set is
$
E=\{(a,b),(a,d),(b,c),(d,e),(c,f),(e,f)\},
$
forming two parallel chains
$
a \rightarrow b \rightarrow c \rightarrow f
$
and
$
a \rightarrow d \rightarrow e \rightarrow f.
$
There is one renewable resource $r$ with capacity $c_1=2$. Each non-dummy activity requires one unit, i.e.,
$
u_{b,1}=u_{c,1}=u_{d,1}=u_{e,1}=1,
$
and
$
u_{a,1}=u_{f,1}=0.
$
The durations are
$
\tau_a=\tau_f=0,\quad \tau_b=3,\quad \tau_c=1,\quad \tau_d=3,\quad \tau_e=2.
$
An optimal schedule has makespan $5$ with
$
S_a=S_b=S_d=0,\quad S_c=S_e=3,\quad S_f=5.
$ Figure~\ref{fig:petri_compare}(a) summarizes the problem instance parameters.
\end{example}

\section{Timed Transition Petri Net with Resources}

A \emph{Petri net}~\cite{petri1966communication} is a bipartite graph consisting of \emph{places} 
(denoted by circles) and \emph{transitions} (rectangles) connected 
by directed arcs. Places hold \emph{tokens}, whose 
distribution defines the \emph{marking} $M$. Formally, 
a Petri net is the tuple $(P, T, F, W, M_0)$, where $P$ 
and $T$ are finite sets of places and transitions, 
$F \subseteq (P \times T) \cup (T \times P)$ is the arc 
set, $W: F \rightarrow \mathbb{Z}^+$ is a weight function, 
and $M_0$ is the initial marking. 
We write $\bullet t = \{p \in P \mid (p,t) \in F\}$ for the 
input places of transition $t$. A transition $t$ is 
\textit{enabled} if every place $p \in \bullet t$ holds at 
least $W(p,t)$ tokens; firing consumes and deposits tokens 
accordingly, thereby modeling concurrency and synchronization. In our formulation, each reachable marking corresponds to a search state, and the reachable markings define the \textit{reachability graph} that our approach generates and searches incrementally.

To distinguish the original RCPSP formulation from its Petri-net encoding,
we use $j \in \mathbb{A}$ for RCPSP activities and $t \in T$ for Petri-net
transitions. Below, %In the construction below, 
each transition corresponds to a single
activity. We extend the classical Petri net to a TTPNR with time and resource perspectives,
\[
\mathcal{N}=(P,T,F,W,M_0,\lambda,\tau_T,R,C),
\]
where $\lambda:T \to \mathbb{A}$ maps each Petri-net transition to its
corresponding RCPSP activity, $\tau_T:T\to\mathbb{N}_0$ assigns a duration to
each transition, and $R,C$ define the resource types and their capacities. Each transition inherits the duration of its associated
activity 
($
\tau_T(t):=\tau_{\lambda(t)}
$). 
A dedicated subset $P_R \subset P$ of resource places enforces capacity
constraints: $M_0$ allocates exactly $c_i$ tokens to each resource place. The
single enabling rule $M(p) \ge W(p,t)$ then guarantees that an activity can
start only if both its precedence conditions and resource requirements are
satisfied. This unified enabling rule is a key advantage over prior RCPSP search formulations, where precedence and resource constraints require separate conflict-resolution mechanisms outside the search space.

\begin{figure*}[t]
\centering

% \resizebox{\textwidth}{!}{%
\begin{subfigure}[t]{0.28\linewidth}
\vspace{0pt}
\centering
\footnotesize
\begin{tabular}{@{}llll@{}}
\multicolumn{4}{l}{\textbf{Problem instance}} \\
\hline
act. & succ. & dur. & $r$ \\
\hline
$a$ & $b$, $d$ & 0 & 0 \\
$b$ & $c$    & 3 & 1 \\
$c$ & $f$    & 1 & 1 \\
$d$ & $e$    & 3 & 1 \\
$e$ & $f$    & 2 & 1 \\
$f$ & —    & 0 & 0 \\
\end{tabular}

\smallskip
\begin{tabular}{@{}ll@{}}
\multicolumn{2}{l}{\textbf{Capacity}} \\
\hline
$r$ & 2 \\
\end{tabular}
\caption{}
\label{fig:instance}
\end{subfigure}
\hfill
\begin{subfigure}[t]{0.31\linewidth}
\vspace{0pt}
    \centering
    \includegraphics[width=\linewidth]{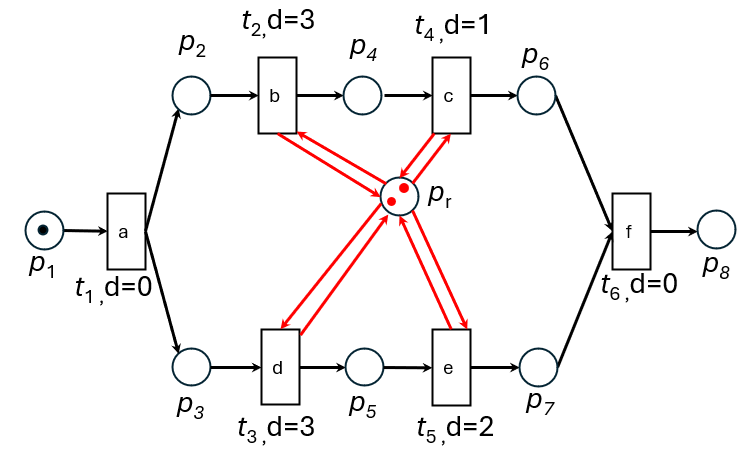}
    \caption{}
    \label{fig:petri1}
\end{subfigure}
\hfill
\begin{subfigure}[t]{0.40\linewidth}
\vspace{0pt}
    \centering
    \includegraphics[width=\linewidth]{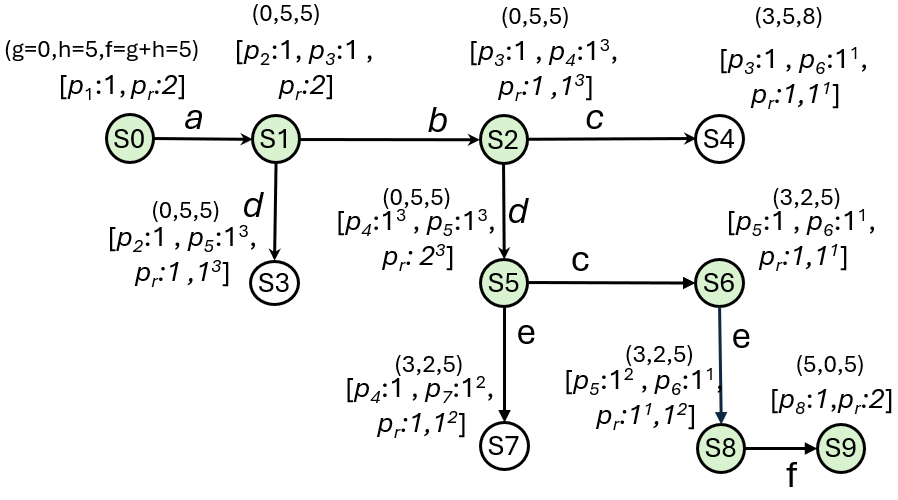}
    \caption{}
    \label{fig:search_graph}
\end{subfigure}%}
\caption{(a) RCPSP instance; (b) TTPNR initial marking; (c) $A^*$ search over the induced reachability graph. 
Nodes show $(g, h, f)$ and token delays $\theta$, 
green nodes are expanded.}
\label{fig:petri_compare}
\end{figure*}

Figure~\ref{fig:petri_compare}(b) illustrates the TTPNR formulation of the RCPSP instance from Example~\ref{exm1} (Figure~\ref{fig:petri_compare}(a)). Here, $a$ and $f$ denote the dummy start and dummy finish activities, respectively. The figure shows the initial marking $M_0$, in which a single token is placed in $p_1$ and two in the resource place $p_r$ (resource type index is omitted for readability). The only enabled transition at $M_0$
is $a$; firing it consumes the token from $p_1$ and produces 
tokens in $p_2$ and $p_3$, making activities $b$ and $d$ 
available to compete for the shared resource.

\section{Modeling RCPSP as a Search Problem} \label{sec:search_RCPSP}
To solve the RCPSP via heuristic search, we induce a reachability 
graph directly from the TTPNR. In this search graph, each node 
represents a system state, 
and each edge corresponds to the commitment to execute a specific 
project activity by firing a Petri-net transition.

%\subsection{State}
A \emph{state} $s \in \mathbb{S}$ is a marking augmented with relative delays on tokens. By encoding time as token delays $\theta$, we avoid a global clock, allowing more states to be recognized as identical and safely pruned; heuristic consistency then guarantees that the first expansion of any state is optimal. Formally,
\begin{align*}
s = \left[ p_1: l_1^{\theta_{l_1}}, \dots, p_n: 
l_n^{\theta_{l_n}}, \,\, p_{r_1}: r_1^{\theta_{r_1}}, 
\dots, p_{r_k}: r_k^{\theta_{r_k}} \right]
\end{align*}
where a token with delay $\theta>0$ is not yet available, while $\theta=0$
(omitted by convention) means immediate availability. Here,
$l_i^{\theta_{l_i}}$ denotes $l_i$ tokens in place $p_i$ with delay
$\theta_{l_i}$, and $r_j^{\theta_{r_j}}$ denotes $r_j$ resource tokens with
delay $\theta_{r_j}$ (only positive token counts are shown).

%\subsection{Start State and Goal State}
The \emph{start state} $s_0$ has one token in the input place of the dummy start transition and all resource places at their initial marking; the \emph{goal state} has one token in the output place of the dummy finish transition with resource places restored.

%\subsection{Actions and Successors}
We consider a transition $t \in T$ \emph{enabled} if its input places contain
sufficient tokens, irrespective of their current relative delays.
Firing $t$ commits the corresponding activity $\lambda(t)$ and proceeds as follows. (1) \emph{Time jump:} Consume the required input tokens, prioritizing tokens with smallest relative delay. The induced time jump is $\Delta t = \max(\theta_{\mathrm{consumed}}).$ (2) \emph{Delay and token generation:} All remaining token delays are reduced by $\Delta t$: $\theta^{\mathrm{new}} = \max(0,\theta-\Delta t).$ Newly produced tokens are assigned the delay $\tau_T(t)=\tau_{\lambda(t)}.$ Also, each node maintains the path cost $g(s') = g(s) + \Delta t$ for use by $A^*$.

%The path cost $g(s') = g(s) + \Delta t$ is maintained at each search node for use by $A^*$.

\iffalse
\begin{enumerate}
    \item \emph{Time jump:} Consume the required input tokens, prioritizing
    tokens with smallest relative delay. The induced time jump is
    $
    \Delta t = \max(\theta_{\mathrm{consumed}}).
    $

    \item \emph{Delay and token generation:} All remaining token delays
    are reduced by $\Delta t$:
    $
    \theta^{\mathrm{new}} = \max(0,\theta-\Delta t).
    $
    Newly produced tokens are assigned the delay
    $
    \tau_T(t)=\tau_{\lambda(t)}.
    $
\end{enumerate}

The path cost $g(s') = g(s) + \Delta t$ 
is maintained at each search node for use by $A^*$.
\fi

\subsection{Heuristics}
We use two lower bounds on the remaining makespan from a state $s$:
the critical-path bound $h_{CP}(s)$, obtained by relaxing resource
capacities, and the resource-load bound $h_{res}(s)$, obtained by
relaxing precedence constraints. Their combination is
$h_{\max}(s)=\max(h_{CP}(s),\,h_{res}(s))$.

\subsubsection{Critical-Path Heuristic}

$h_{CP}(s)$ is the length of the longest \emph{residual} precedence path to the dummy finish activity $n{+}1$, computed by propagating \emph{residual} earliest finish times, $REF$, in topological order:
\[
REF_j =
\begin{cases}
0 & \text{if $j$ is completed,}\\
\theta_j & \text{if $j$ is currently executing,}\\
\displaystyle\max_{i\in \bullet j} REF_i + \tau_j & \text{if $j$ has not yet started,}
\end{cases}
\]
initialized with $REF_0=0$, and $h_{CP}(s)=REF_{n+1}$.

\textbf{Proposition 1.} $h_{CP}$ is admissible and consistent.

\begin{proof}
\textit{Admissibility.} Removing resource constraints cannot extend %only shorten
the remaining completion time, so $h_{CP}(s)\le h^*(s)$.

\textit{Consistency.} Consider a successor $s'$ of $s$ with time
increment $\delta(s,s'):=g(s')-g(s)$. For every activity $j$, its
residual duration decreases by at most $\delta(s,s')$: executing
activities lose at most $\delta(s,s')$, unstarted activities are
unchanged, and a newly started activity $\lambda(t)$ receives residual
delay $\tau_{\lambda(t)}$ in $s'$, identical to its value before
firing. Since precedence enforces sequential execution along any single
path, at most one activity on the longest path $P^*$ is executing in
$s$, so the total residual length of $P^*$ decreases by at most
$\delta(s,s')$. Since $h_{CP}(s')$ is the maximum over all paths,
\[
h_{CP}(s')\ge h_{CP}(s)-\delta(s,s'),
\]
which rearranges to $h_{CP}(s)\le c(s,s')+h_{CP}(s')$.
\end{proof}

\subsubsection{Resource-Based Heuristic} 

$h_{res}(s)$ is the minimum time to clear the remaining workload on the most heavily loaded resource, ignoring precedence constraints. Let $\mathrm{active}(s)$ and $\mathrm{unstarted}(s)$ denote the
currently executing and not-yet-started activities, respectively.
\[
h_{res}(s)=
\max_{r\in R}
\frac{
\displaystyle\sum_{a\in \mathrm{active}(s)} \theta_a\, u_{a,r}
+
\sum_{a\in \mathrm{unstarted}(s)} \tau_a\, u_{a,r}
}{c_r}.
\]

\textbf{Proposition 2.} $h_{res}$ is admissible and consistent.

\begin{proof}
\textit{Admissibility.} Dropping precedence constraints can only reduce
the remaining makespan, so $h_{res}(s)\le h^*(s)$.

\textit{Consistency.} The argument mirrors that of $h_{CP}$: firing $t$ advances the clock by $\delta(s,s')$, reducing the residual workload numerator by at most $\delta(s,s')\cdot c_r$ per resource (executing activities lose at most $\delta(s,s')\cdot u_{a,r}\le \delta(s,s')\cdot c_r$, unstarted activities are unchanged, and
$\lambda(t)$ satisfies $\theta_{\lambda(t)}=\tau_{\lambda(t)}$
immediately after firing). 
Dividing by $c_r$ and taking the maximum over resources yields $h_{res}(s)\le c(s,s')+h_{res}(s')$.
\end{proof}

%\subsubsection{Combined Heuristic}
%\subsubsection{Combined Heuristic and Caching}
%$h_{\max}(s)=\max(h_{CP}(s),h_{res}(s))$.

\textbf{Proposition 3.} A combined heuristic $$h_{\max}(s)=\max(h_{CP}(s),h_{res}(s))$$ is admissible and consistent.
\begin{proof}
The max over admissible and consistent heuristics is admissible and consistent~\cite{russell2020artificial}.
\end{proof}

%By Propositions~1 and~2, $h_{CP}$ and $h_{res}$ are consistent, and hence $h_{\max}$ is consistent. as well. 
Consistency guarantees that %graph-search 
$A^*$ remains optimal while eliminating the need to reopen expanded states~\cite{hart1968formal,10.1145/3828.3830}. Our formulation also enables an effective caching optimization: whenever $\delta(s,s')=0$, both lower bounds are preserved, so $h_{\max}(s')=h_{\max}(s)$. Thus, zero-cost successors can reuse the parent's cached heuristic value, without recomputation.

In the resulting $A^*$ search, we prioritize nodes with minimal $f(s)=g(s)+h_{\max}(s)$. $f$-ties are broken by preferring higher $g$-cost, then more finished activities, then more active activities, then FIFO, driving the search depth-first through equal $f$-cost plateaus.

\begin{table*}[t!] % The asterisk (*) tells LaTeX to span both columns
\centering
 %\resizebox{0.9\textwidth}{!}{% illigal command
\small
\begin{tabular}{l | r r r | r r r | r r r}
\hline
{\textbf{Metric / Parameter}} & \multicolumn{3}{c|}{\textbf{J30}} & \multicolumn{3}{c|}{\textbf{J60}} & \multicolumn{3}{c}{\textbf{J90}} \\

& \textbf{TTPNR} & \textbf{CBC} & \textbf{SCIP} & \textbf{TTPNR} & \textbf{CBC} & \textbf{SCIP} & \textbf{TTPNR} & \textbf{CBC} & \textbf{SCIP} \\
\hline
\multicolumn{10}{c}{\textit{Overall Performance}} \\
\hline
\textbf{Success Rate (\%)} 
    & \textit{(+27.5)} \textbf{98.96}& 47.29 & 71.46 
    & \textit{(+19.6)} \textbf{47.71}  & 28.13 &  8.75 
    & \textit{(+23.1)} \textbf{36.04}  & 12.92 &  0.21 \\
\textbf{Avg. Time (s)} 
    & \textit{(-78.8\%)} \textbf{7.97}  & 42.14 & 37.65 
    & \textit{(-77.5\%)} \textbf{19.33}  & 85.96 & 186.42 
    & \textit{(-96.2\%)} \textbf{5.97}  & 197.37 & 155.97 \\
\hline
\multicolumn{10}{c}{\textit{Success Rate (\%) by Resource Strength (RS)}} \\
\hline
\textbf{RS = 1.0} &\textbf{100.00} & 95.00 & \textbf{100.00} & \textbf{94.17} & 56.67 & 27.50 & \textbf{87.50} & 30.83 & 0.83 \\
\textbf{RS = 0.7} &  \textbf{98.33} & 60.83 &  95.83 & \textbf{71.67} & 46.67 &  5.00 & \textbf{43.33} & 18.33 & 0.00 \\
\textbf{RS = 0.5} &  \textbf{99.17} & 24.17 &  68.33 & \textbf{22.50} &  9.17 &  1.67 & \textbf{13.33} &  2.50 & 0.00 \\
\textbf{RS = 0.2} &  \textbf{98.33} &  9.17 &  22.50 &  \textbf{2.50} &  0.00 &  0.83 &  0.00 &  0.00 & 0.00 \\
\hline
\multicolumn{10}{c}{\textit{Success Rate (\%) by Resource Factor (RF)}} \\
\hline
\textbf{RF = 1.0} &\textbf{100.00} & 38.33 & 51.67 & \textbf{49.17} & 35.83 &  3.33 & \textbf{40.00} & 15.00 & 0.83 \\
\textbf{RF = 0.75} & \textbf{100.00} & 38.33 & 65.00 & \textbf{49.17} & 40.83 &  5.83 & \textbf{34.17} & 15.83 & 0.00 \\
\textbf{RF = 0.50} &  \textbf{98.33} & 44.17 & 72.50 & \textbf{50.83} & 30.00 &  5.83 & \textbf{33.33} & 19.17 & 0.00 \\
\textbf{RF = 0.25} &  \textbf{97.50} & 68.33 & \textbf{97.50} & \textbf{41.67} &  5.83 & 20.00 & \textbf{36.67} &  1.67 & 0.00 \\
\hline
\end{tabular}%}
\caption{%Comprehensive performance comparison: 
Overall Success Rate and Average Time, alongside breakdowns by Resource Strength (RS) and Resource Factor (RF).}
\label{tab:comprehensive_results_wide}
\end{table*}

Figure~\ref{fig:petri_compare}(c) illustrates the $A^*$ 
guided search  over the induced reachability 
graph of Example~\ref{exm1}. Each node shows $(g, h, f)$ values and active token delays $\theta$. 
At each expansion, a successor is generated for enabled transitions, and the prioritized node is selected for expansion. 
% Green nodes are those expanded by the search.

\section{MIP Formulation} \label{sec:MIP_RCPSP}

Since MIP-based branch-and-cut is the dominant exact approach for RCPSP~\cite{dorndorf2000branch}, we compare against two strong off-the-shelf branch-and-cut solvers, SCIP~\cite{achterberg2009scip} and CBC~\cite{cbc2005}, using the standard time-indexed formulation~\cite{artigues2015mixed}: a binary variable $x_{jt} \in \{0,1\}$ indicates whether activity $j$ starts at time $t$. The full encoding is:
 \begin{alignat}{2} &\min & \sum_{t \in H}\ & t \cdot x_{n+1,t} \\ &\text{s.t.} & \sum_{t \in H} x_{jt} &= 1 \quad \forall j \in \mathbb{A} \label{eq:once}\\ & & S_j + \tau_j &\leq S_i \quad \forall (j,i) \in E \label{eq:prec}\\ & & \sum_{j \in \mathbb{A}} u_{j,r} \sum_{q=t-\tau_j+1}^{t} x_{jq} &\leq c_r \quad \forall r \in R,\ \forall t \in H \label{eq:res} \end{alignat}
Constraints~(\ref{eq:once})--(\ref{eq:res}) ensure activities start exactly once, enforce precedence and 
resource capacity, with $S_j=\sum_{t\in H} t\,x_{jt}$ and $H=\{0,\dots,T\}$, where 
$T=\sum_{j\in\mathbb{A}} \tau_j$.

\section{Experiments and Discussion}
We evaluated $A^*$ (denoted TTPNR) against SCIP and CBC on the standard public RCPSP benchmark suite PSPLIB~\citep{kolisch1997psplib}, using the J30, J60, and J90 sets (30, 60, and 90 activities; 480 instances per set). Each instance has four renewable resource types, and the instances in each set are organized into 48 groups of 10 with shared generation parameters. These parameters (discussed later) vary structural properties and resource tightness, yielding a diverse collection of challenging RCPSP instances.
Evaluations were run sequentially on a single-core Intel Xeon Gold 6248R CPU @ 3.00GHz with a 5-minute timeout. To ensure a rigorous comparison of the underlying search engines and branching mechanics, the MIP solvers and TTPNR  were executed without preprocessing. TTPNR was implemented in C++ and will be publicly available.

%TTPNR operates directly on the underlying search problem without state-space reduction techniques.

Table~\ref{tab:comprehensive_results_wide} summarizes success rate, average time, and performance breakdowns by problem features. TTPNR outperforms both MIP solvers in success rate and is substantially faster across all datasets.
\subsection{Impact of Problem Features on TTPNR}
\emph{Network Complexity (NC)} measures how strongly precedence constraints restrict activity order, counting only non-redundant arcs (redundant arcs are implied by transitivity). Higher NC makes the schedule more sequential, reducing the number of simultaneously enabled activities, reducing the branching factor. From our experiments, NC proved to be the least impactful parameter for both MIP solvers and TTPNR, and is thus omitted from Table~\ref{tab:comprehensive_results_wide}.

\emph{Resource Strength (RS)} measures how restrictive the resource capacities are. At RS = 1, resources are abundant enough that 
competition for resources is minimal, while RS$~=0$ implies that capacities are near their minimum feasible levels and competition for resources is maximal. When RS is low, many delays are caused by resource conflicts rather than precedence, so the true remaining makespan can be much larger than the precedence-only lower bound. This makes $h_{CP}$, and therefore $h_{\max}$, less informative for search, weakening $A^*$ guidance and reducing success rates (RS section of Table~\ref{tab:comprehensive_results_wide}).

\emph{Resource Factor (RF)} measures the average proportion of resource types required per activity. While SCIP strictly degrades as RF increases, CBC exhibits mixed sensitivity. TTPNR demonstrates robustness to RF variations. In TTPNR multi-resource requirements involve transitions concurrently consuming tokens from multiple places, preventing branching factor inflation and ensuring stable success rates for any RF (RF section of Table~\ref{tab:comprehensive_results_wide}).

\section{Conclusion}
We formulated RCPSP as optimal search on the reachability graph of a TTPNR with relative-delay tokens, and solved it with $A^*$ guided by a consistent heuristic combining Critical Path and resource-based lower bounds. More broadly, any scheduling problem expressible as a timed Petri net with resources inherits the same search machinery, including consistent heuristics, safe duplicate pruning, and zero-cost caching.
Empirically, heuristic search and MIP degrade along independent axes---resource tightness for $A^*$, formulation size for MIP---with resource strength mediating which solver benefits from scale.
Future work includes stronger resource-aware heuristics for low-resource-strength instances, algorithm selection across the two paradigms, and bidirectional search methods such as BAE*~\cite{sadhukhan2013bidirectional}, which fit our state representation.

\bibliography{aaai2026}


\end{document}